\documentclass[sigconf]{acmart}

\usepackage{amsmath}
\usepackage{booktabs}
\usepackage{listings}
\usepackage{algorithm}
\usepackage{algorithmic}
\usepackage{xcolor}
\usepackage{enumitem}

\setcopyright{none}
\renewcommand\footnotetextcopyrightpermission[1]{}
\acmConference[Preprint]{arXiv Preprint}{February 2026}{}
\acmYear{2026}
\acmISBN{}
\acmDOI{}

\newcommand{\hrepo}[1]{h^{\text{repo}}_{#1}}

\begin{document}

\title{Hierarchical Embedding Fusion for Retrieval-Augmented Code Generation}

\author{Nikita Sorokin}
\authornote{Corresponding author.}
\affiliation{%
  \institution{MWS AI}
  \city{Moscow}
  \country{Russia}}
\email{n.sorokin@mts.ai}

\author{Ivan Sedykh}
\affiliation{%
  \institution{MWS AI}
  \city{Moscow}
  \country{Russia}}
\email{i.sedykh@mts.ai}

\author{Valentin Malykh}
\affiliation{%
  \institution{MWS AI}
  \city{Moscow}
  \country{Russia}}
\affiliation{%
  \institution{ITMO University}
  \city{Saint Petersburg}
  \country{Russia}}
\email{v.malykh@mts.ai}

\begin{abstract}
Retrieval-augmented code generation commonly conditions a decoder on large retrieved snippets, which couples online cost to repository size and introduces long-context noise. We present \emph{Hierarchical Embedding Fusion (HEF)}, a two-stage repository representation for code completion: (i) an offline cache that compresses repository chunks into a reusable hierarchy of dense vectors using a small fuser model, and (ii) an online interface that maps a small number of retrieved vectors into learned pseudo-tokens consumed by a code generator. This replaces thousands of retrieved tokens with a fixed pseudo-token budget while retaining access to repository-level information.

On RepoBench and RepoEval, HEF with a 1.8B pipeline attains comparable exact-match accuracy to snippet-based retrieval baselines while operating at sub-second median latency on a single A100. Compared to graph- and iterative-retrieval systems in our setup, HEF reduces median end-to-end latency by \(13\!\times\)--\(26\!\times\). We additionally introduce a \emph{utility-weighted likelihood} signal to filter training contexts and report ablations over pseudo-token budget, embedding models, and robustness to harmful retrieval. These results support hierarchical dense caching as an effective mechanism for low-latency repository-aware code completion.
\end{abstract}



\maketitle

\begin{figure}[t]
\centering
\includegraphics[width=0.6\columnwidth]{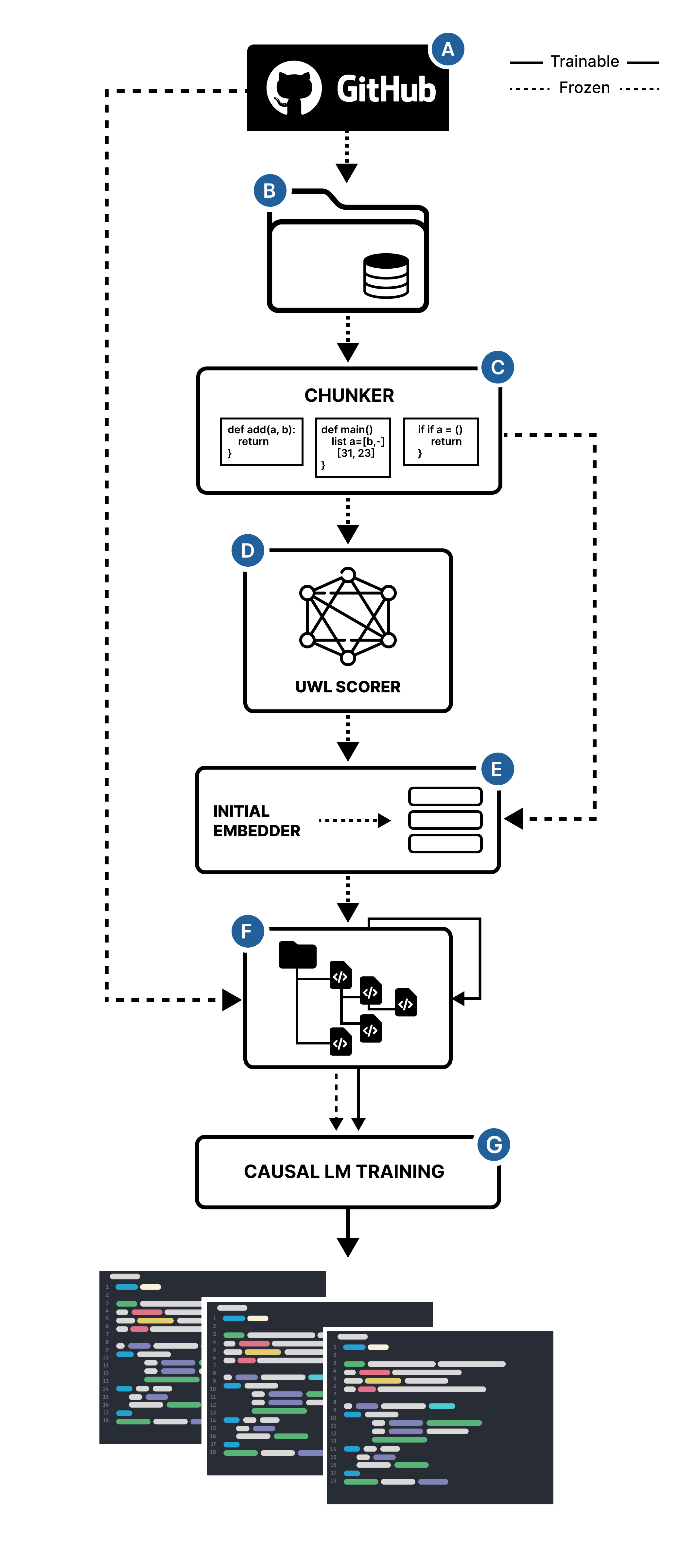}
\caption{HEF pipeline overview. \textbf{(A)} GitHub repositories are crawled and filtered. \textbf{(B)} Repositories are deduplicated and stored. \textbf{(C)} Files are split into $\le$512-token chunks. \textbf{(D)} UWL scoring filters training contexts by utility. \textbf{(E)} Frozen embedder maps chunks to dense vectors. \textbf{(F)} Fuser recursively merges vectors into file/module/repo hierarchy (cached). \textbf{(G)} At inference, retrieved vectors become pseudo-tokens for the generator.}
\Description{A diagram showing the two-stage HEF pipeline: offline hierarchical fusion of code chunks into cached vectors, and online pseudo-token conditioning of the code generator.}
\label{fig:pipeline}
\end{figure}

\begin{figure}[t]
\centering
\includegraphics[width=0.95\columnwidth]{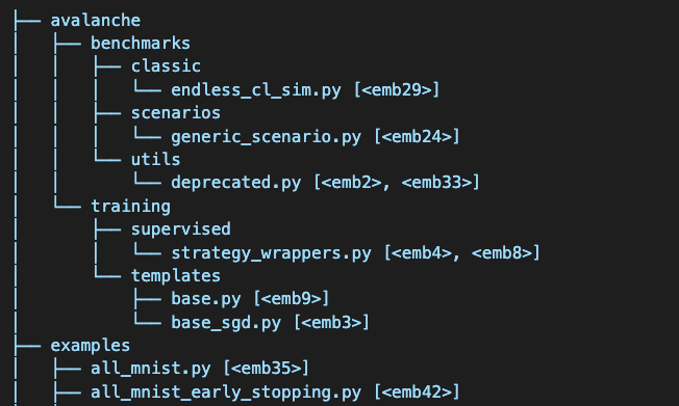}
\caption{Example of a fused repository tree: chunks merge into file embeddings, which merge into module and repository embeddings.}
\Description{A tree diagram showing how code chunks are hierarchically fused into file-level, module-level, and repository-level embeddings.}
\label{fig:example}
\end{figure}
\section{Introduction}

Repository-level code completion requires predicting the next line (or lines) of code given both an in-file prefix and cross-file context from the surrounding project. Unlike single-file completion, the generator must resolve symbols defined elsewhere, respect project-wide APIs, and follow architectural patterns established across modules. In practice this means the model needs access to imported classes, function signatures from other files, and type definitions that may live several directories away. Without such cues, even capable language models hallucinate identifiers or produce code that is syntactically valid but semantically inconsistent with the repository.

Retrieval-augmented code completion (RACC) systems which retrieve cross-file context to condition a code model can be roughly grouped into three common patterns.
\emph{Snippet-injection} methods retrieve raw code chunks and concatenate them into the prompt; while simple, they couple online latency to the number of retrieved tokens and introduce noise when irrelevant fragments enter the context window.
\emph{Structure-aware} approaches including graph-based and iterative-retrieval systems improve relevance by exploiting data-flow or control-flow information, but require multiple model calls or expensive graph traversals at query time.
\emph{Dense-caching} techniques compress context into continuous representations, yet such methods have rarely been systematically evaluated for repository-level code completion with a pseudo-token interface that fully replaces raw snippets.

We present \textbf{Hierarchical Embedding Fusion (HEF)}, a two-stage approach to repository-aware code completion (Figure~\ref{fig:pipeline}).
\emph{Offline}, HEF compresses each repository into a reusable hierarchy of dense vectors: code chunks are embedded, then recursively merged by a small ``Fuser'' model into file-, module-, and repository-level representations that can be cached and updated incrementally.
\emph{Online}, a fixed number of hierarchy nodes are retrieved and mapped to learned pseudo-tokens that condition the code generator, replacing thousands of raw context tokens with a bounded pseudo-token budget.
Because the heavy lifting happens once per repository, HEF is compatible with any structure-aware retrieval front-end while keeping per-query cost low.

Several components underlying our approach have been studied separately. Long-context Transformers reduce the cost of processing lengthy inputs via recurrence or sparse attention~\cite{dai2019transformerxlattentivelanguagemodels, rae2019compressivetransformerslongrangesequence, beltagy2020longformer}. Dense retrieval and contrastive embedding learning provide the standard machinery for selecting relevant context~\cite{karpukhin-etal-2020-dense, gao-etal-2021-simcse, reimers2019sentencebertsentenceembeddingsusing}. Learned pseudo-token interfaces are well-studied under soft prompt and prefix tuning~\cite{li2021prefixtuning, lester2021prompttuning, liu2022ptuningv2}. However, these techniques have not been combined into an end-to-end repository-level pipeline with hierarchical dense caching and a pseudo-token interface for code completion.

Our contribution is the \emph{end-to-end} integration of these components: (i) a concrete pipeline that starts with a strong code embedder, builds a multi-resolution \emph{hierarchical} dense cache via a lightweight fuser, and exposes retrieved nodes to a causal generator through a bounded pseudo-token interface; and (ii) an \emph{unsupervised} data construction procedure that derives training signals from raw repositories without requiring labelled query--context alignments. Together with two training regimes and an interactive-latency evaluation protocol, this provides a practical recipe for replacing snippet injection with compact fused representations in repository-level completion.

Our contributions are:
\begin{itemize}[leftmargin=1.5em,itemsep=2pt]
  \item \textbf{Method.} A hierarchical dense cache plus a pseudo-token interface that decouples repository size from online prompt length.
  \item \textbf{Two training regimes} --- contrastive Fuser pre-training and joint end-to-end optimization---with analysis of what each learns and when to prefer one over the other.
  \item \textbf{Accuracy and latency measurements}  on RepoBench and RepoEval, ablations over pseudo-token budget and embedding backbones, and robustness tests against harmful retrieved context.
\end{itemize}
On these benchmarks, HEF achieves comparable exact-match accuracy to snippet-based methods while operating at sub-second median latency on a single A100.

\section{Related Work}
\label{sec:related}

\subsection{Repository-Level Code Completion}
Repository-level code completion extends single-file prediction to the setting where cross-file context---imported classes, shared type definitions, and project-wide APIs---is required for accurate generation. RepoBench~\cite{liu-etal-2024-repobench} formalises this task and provides benchmarks that separately evaluate retrieval, completion, and end-to-end pipelines. RepoCoder~\cite{zhang-etal-2023-repoeval} introduces RepoEval with unit-test-based evaluation and proposes an iterative retrieve-and-generate loop. Repoformer~\cite{wu2024repoformer} studies selective retrieval, using self-supervised signals to decide when retrieval is beneficial and avoid unnecessary retrieval overhead. These benchmarks and baselines establish that naively concatenating retrieved snippets trades improved relevance for increased prompt length and online latency.

\subsection{Snippet-Injection RAG for Code}
The simplest retrieval-augmented approach retrieves raw code chunks and concatenates them into the prompt, following the standard retrieve-then-condition paradigm established in RAG-style generation~\cite{ borgeaud2022retro} ~\cite{lewis2020rag, guu2020realm}. In code settings, this snippet-injection pattern has been widely adopted by retrieval-augmented code generation systems, which supply retrieved code or documentation as additional prompt context~\cite{parvez2021redcoder,shrivastava2023repofusion,wang2025coderagbench}. While straightforward, this strategy couples online cost to the number of retrieved tokens and introduces noise when irrelevant fragments enter the context window. Scaling pretrained code models such as Codex~\cite{chen-etal-2021-evaluating} and Qwen 2.5-Coder~\cite{hui2024qwen25codertechnicalreport} improves generation quality, yet the fundamental latency-relevance trade-off of raw snippet injection remains.

\subsection{Structure-Aware Retrieval}
Graph-based methods exploit program structure to improve retrieval relevance. DRACO~\cite{cheng2024draco} builds a repository-specific context graph using extended dataflow analysis and retrieves background knowledge from this graph to construct prompts. GraphCoder~\cite{liu2024graphcoder} constructs a code context graph capturing control and data dependencies and applies coarse-to-fine retrieval to locate relevant snippets. These approaches improve precision via program structure, but require graph construction and online graph traversals for each query, creating latency and memory overhead on large or frequently changing codebases.

\subsection{Iterative and Adaptive Retrieval}
Iterative retrieval systems dynamically adjust context during generation. RLCoder~\cite{wang2024rlcoder} formulates retrieval as a reinforcement-learning problem: a dual-encoder iteratively adds chunks until the estimated utility stalls. Dynamic RAG~\cite{shapkin2024drag} replaces long retrieved passages with learnable vocabulary tokens that the generator can expand on demand, though it treats functions in isolation. These approaches improve relevance through adaptivity at the cost of multiple model calls or specialised stopping heuristics.

\subsection{Pseudo-Token Conditioning and Soft Prompts}
Learned continuous prompts provide an alternative interface between retrieved information and generation. Prefix-tuning~\cite{li2021prefixtuning} learns continuous ``virtual tokens'' that condition a frozen LM, providing an early template for pseudo-token interfaces. Prompt tuning~\cite{lester2021prompttuning} learns soft prompt embeddings prepended to inputs, showing that such interfaces become competitive at scale. P-Tuning v2~\cite{liu2022ptuningv2} extends soft prompts by injecting continuous prompts more deeply into the transformer, improving capacity and robustness. These methods demonstrate that learned token-like representations can effectively condition generation, though they have been less systematically instantiated as repository-level caching pipelines for code completion.

\subsection{Dense Retrieval and Representation Learning}
Dense passage retrieval~\cite{karpukhin-etal-2020-dense} popularised dual-encoder retrieval trained end-to-end with contrastive objectives, establishing a standard template for embedding-based retrieval. Contrastive embedding methods such as SimCSE~\cite{gao-etal-2021-simcse} and Sentence-BERT~\cite{reimers2019sentencebertsentenceembeddingsusing} produce similarity-friendly dense spaces supporting cosine-based retrieval. Retrieval-augmented LMs like RETRO~\cite{borgeaud2022retro} demonstrate how external embedding stores can complement parametric models, motivating cache-style augmentation strategies.

\subsection{Long-Context Modeling}
Long-context Transformers reduce the cost of processing lengthy inputs. Transformer-XL~\cite{dai2019transformerxlattentivelanguagemodels} introduces recurrence across segments to extend effective context beyond fixed windows. Sparse or approximate attention variants such as Longformer~\cite{beltagy2020longformer} and BigBird~\cite{zaheer2020bigbird} reduce long-sequence cost via structured sparsity. Compressive Transformers~\cite{rae2019compressivetransformerslongrangesequence} compress past activations into a fixed-size memory. These techniques address the cost of long inputs but do not inherently provide a retrieval interface for selecting which context to include.

\subsection{Fusion-in-Decoder Approaches}
RepoFusion~\cite{shrivastava2023repofusion} concatenates the full repository into the decoder's context during pre-training, eliminating explicit retrieval at inference. This grants strong global awareness but forces the decoder to attend to thousands of tokens for every query, making inference costly regardless of how much context is actually relevant.

\section{Method}
\label{sec:method}

\subsection{Task Definition and Notation}
\label{sec:notation}

We define repository-level code completion as follows. A \textbf{repository} $\mathcal{R} = \{c_1, \dots, c_N\}$ consists of $N$ code chunks, where each chunk $c_i$ is a semantically coherent unit of source code (e.g., a function, method, or small class) containing up to 512 tokens.

Given an \textbf{in-file prefix} $x$ (the code preceding the cursor), the task is to predict the \textbf{gold completion} $y^*$ (the ground-truth continuation). A \textbf{generator} $G_\phi$ with parameters $\phi$ produces the output: $\hat{y} = G_\phi(x, \mathcal{P})$, where $\mathcal{P}$ represents the conditioning context.

In classical retrieval-augmented generation, $\mathcal{P}$ is raw text from the $K$ nearest chunks, so prompt length scales with retrieval budget. HEF replaces raw text with a fixed set of $m$ \textbf{pseudo-tokens} $\mathcal{P} = \{p_1, \dots, p_m\}$, continuous vectors that condition the generator without expanding the discrete token sequence.

The HEF pipeline involves:
\begin{itemize}[leftmargin=1.5em,itemsep=1pt]
  \item \textbf{Embedder} $E$: maps chunks to vectors $h_i = E(c_i) \in \mathbb{R}^d$
  \item \textbf{Fuser} $F_\theta$: merges child vectors into parent vectors, building hierarchy $\mathcal{H}_{\mathcal{R}}$
  \item \textbf{Retriever}: given query $q$, returns top-$K$ nodes $\mathcal{Z} = \textsc{Retrieve}(q, \mathcal{H}_{\mathcal{R}})$
  \item \textbf{Projector} $\pi_\psi$: maps retrieved vectors to pseudo-tokens $\mathcal{P} = \pi_\psi(\mathcal{Z})$
  \item \textbf{Generator} $G_\phi$: produces completion conditioned on prefix and pseudo-tokens
\end{itemize}

\subsection{Offline Stage: Hierarchical Cache Construction}
\label{sec:offline}

The offline stage processes each repository once to build a reusable dense cache. Algorithm~\ref{alg:buildcache} details the procedure.

\begin{algorithm}[t]
\caption{\textsc{BuildCache}: Offline Hierarchy Construction}
\label{alg:buildcache}
\begin{algorithmic}[1]
\REQUIRE Repository $\mathcal{R}$, embedder $E$, fuser $F_\theta$, branching factor $b$
\ENSURE Hierarchy $\mathcal{H}_{\mathcal{R}}$ with node-to-span mappings
\STATE $\mathcal{R} \gets \textsc{Chunk}(\mathcal{R})$ \COMMENT{Split files into $\le$512-token chunks}
\FOR{each chunk $c_i \in \mathcal{R}$}
    \STATE $h_i^{(0)} \gets E(c_i)$ \COMMENT{Embed with frozen encoder}
\ENDFOR
\STATE $\ell \gets 0$
\WHILE{$|\{h^{(\ell)}\}| > 1$}
    \STATE Group nodes by file path, then by directory
    \FOR{each group $\mathcal{G} = \{h_1^{(\ell)}, \dots, h_b^{(\ell)}\}$}
        \STATE $h^{(\ell+1)} \gets F_\theta(h_1^{(\ell)}, \dots, h_b^{(\ell)})$ \COMMENT{Fuse children}
        \STATE Store mapping: $h^{(\ell+1)} \to \text{spans of children}$
    \ENDFOR
    \STATE $\ell \gets \ell + 1$
\ENDWHILE
\STATE Index all nodes $\{h^{(\ell)}\}_{\ell=0}^{L}$ with HNSW
\RETURN $\mathcal{H}_{\mathcal{R}}$
\end{algorithmic}
\end{algorithm}

\paragraph{Chunking.} Each source file is split into chunks of at most 512 tokens using lightweight AST-aware slicing. We prefer semantic boundaries (function/class definitions) but fall back to syntactic heuristics (blank lines, indentation changes) when AST parsing fails.

\paragraph{Embedding.} A frozen encoder $E$ (\texttt{Qwen3-Embedding-8B}, output dimension $d{=}4096$) maps each chunk $c_i$ to a dense vector $h_i^{(0)} \in \mathbb{R}^d$. The encoder is never fine-tuned, ensuring embeddings remain stable across cache updates.

\paragraph{Hierarchy construction.} Chunks are grouped by file, then by directory, mirroring the repository's structure. At each level $\ell$, groups of up to $b$ sibling vectors ($b{=}8$) are fused into a single parent vector at level $\ell{+}1$. This produces file-level, module-level, and repository-level representations. Each node stores metadata mapping it back to the source spans it summarises.

\subsection{Fuser Architecture}
\label{sec:fuser}

The \textbf{Fuser} $F_\theta$ is a small causal transformer (\texttt{Qwen-2.5-Coder-0.5B}) that compresses multiple child vectors into one parent vector during offline cache construction. We use a compact model because the fuser is invoked many times (once per internal hierarchy node), and its role is compression rather than generation.

\paragraph{Input representation.} Given $b$ child vectors $\{h_1, \dots, h_b\} \subset \mathbb{R}^d$, we project each to the transformer's hidden dimension $d_f{=}896$ via a learned linear layer: $\tilde{h}_i = W_{\text{in}} h_i + b_{\text{in}}$, where $W_{\text{in}} \in \mathbb{R}^{d_f \times d}$. Each projected vector becomes one input token.

\paragraph{Fusion.} The transformer processes the sequence $[\tilde{h}_1, \dots, \tilde{h}_b]$ with causal attention. We take the final hidden state (position $b$) and project it back: $h^{(\ell+1)} = W_{\text{out}} \cdot \text{Transformer}([\tilde{h}_1, \dots, \tilde{h}_b])_b$, where $W_{\text{out}} \in \mathbb{R}^{d \times d_f}$.

\paragraph{Why a small model?} The quality bottleneck is the downstream generator, not the fuser. Increasing fuser size beyond 0.5B yields diminishing returns as we show later in the ablation study while substantially increasing cache-build time. The fuser operates purely on dense vectors and does not see raw text.

\paragraph{Note on terminology.} The fuser \emph{compresses} during cache construction ($b$ vectors $\to$ 1 vector). At inference, the \emph{projector} (Section~\ref{sec:pseudotokens}) \emph{transforms} each of the $K$ retrieved vectors into a pseudo-token by changing its dimension from $d$ to $d_g$. The fuser reduces count; the projector changes representation space.

\subsection{Online Stage: Query Processing}
\label{sec:online}

At inference time, HEF recieves the query, retrieves relevant hierarchy nodes and converts them to pseudo-tokens. Algorithm~\ref{alg:querycache} details the procedure.

\begin{algorithm}[t]
\caption{\textsc{QueryCache}: Online Completion}
\label{alg:querycache}
\begin{algorithmic}[1]
\REQUIRE Prefix $x$, hierarchy $\mathcal{H}_{\mathcal{R}}$, embedder $E$, projector $\pi_\psi$, generator $G_\phi$
\ENSURE Completion $\hat{y}$
\STATE $q \gets E(\textsc{TruncateLast}(x, 512))$ \COMMENT{Embed query from prefix suffix}
\STATE $\mathcal{Z} \gets \textsc{HNSW-Search}(q, \mathcal{H}_{\mathcal{R}}, K)$ \COMMENT{Retrieve top-$K$ nodes}
\STATE $\mathcal{P} \gets \pi_\psi(\mathcal{Z})$ \COMMENT{Project to $m$ pseudo-tokens}
\STATE $\hat{y} \gets G_\phi(x, \mathcal{P})$ \COMMENT{Generate with pseudo-token conditioning}
\RETURN $\hat{y}$
\end{algorithmic}
\end{algorithm}

\paragraph{Query formation.} We form the query $q$ by embedding the last 512 tokens of the prefix $x$ using the frozen encoder $E$.

\paragraph{Retrieval.} We perform approximate nearest-neighbor search over all hierarchy nodes using HNSW with cosine similarity. Retrieval returns the top-$K$ nodes ($K{=}32$), which may come from any hierarchy level.

\subsection{Pseudo-Token Interface}
\label{sec:pseudotokens}

The \textbf{projector} $\pi_\psi$ converts each of the $K$ retrieved vectors (in $\mathbb{R}^d$) into a pseudo-token (in $\mathbb{R}^{d_g}$, where $d_g$ is the generator's embedding dimension). The number of pseudo-tokens equals the number of retrieved vectors ($m = K$); only the dimension changes.

\paragraph{Architecture.} Each retrieved vector is independently projected via a two-layer MLP with GELU activation:
\[
  p_i = \pi_\psi(z_i) = W_2 \cdot \text{GELU}(W_1 \cdot z_i + b_1) + b_2,
\]
where $W_1 \in \mathbb{R}^{d_{\text{hidden}} \times d}$ and $W_2 \in \mathbb{R}^{d_g \times d_{\text{hidden}}}$. This maps each $z_i \in \mathbb{R}^d$ to $p_i \in \mathbb{R}^{d_g}$.

\paragraph{Injection.} Pseudo-tokens are \textbf{prepended} to the input embedding sequence (soft-prompt style). If the tokenized prefix produces embeddings $[e_1, \dots, e_T]$, the generator receives $[p_1, \dots, p_K, e_1, \dots, e_T]$.

\paragraph{Budget.} We use $K{=}32$ retrieved vectors, yielding 32 pseudo-tokens. Ablations (Section~\ref{sec:ablation}) show 30 to 40 tokens capture most repository-level information; beyond 60 tokens, performance slightly degrades.

\subsection{Training Regimes}
\label{sec:training}

We study two training regimes, summarised in Table~\ref{tab:training_regimes}.

\begin{table}[t]
\caption{Training regimes: what is optimised vs.\ frozen.}
\label{tab:training_regimes}
\small
\begin{tabular}{l l l l}
\toprule
\textbf{Regime} & \textbf{Optimises} & \textbf{Freezes} & \textbf{Objective} \\
\midrule
Contrastive & $F_\theta$ & $E$, $G_\phi$ & $\mathcal{L}_{\text{CL}}$ \\
(pretrain) & (fuser) & & (InfoNCE) \\
\midrule
Separate & $\pi_\psi$, $G_\phi$ & $E$, $F_\theta$ & $\mathcal{L}_{\text{CLM}}$ \\
(fine-tune) & (proj., gen.) & & (LM loss) \\
\midrule
End-to-End & $F_\theta$, $\pi_\psi$, $G_\phi$ & $E$ & $\mathcal{L}_{\text{CLM}}$ \\
& (all except emb.) & & (LM loss) \\
\bottomrule
\end{tabular}
\end{table}

\paragraph{Contrastive pretraining.} We first train the fuser $F_\theta$ with a contrastive objective. For a query chunk $q$ from repository $\mathcal{R}$, the positive is the fused repository vector $h^{\text{repo}}_+$ and negatives are repository vectors from other mini-batch samples:
\begin{equation}
\mathcal{L}_{\mathrm{CL}} = -\log
  \frac{\exp(s(q, \hrepo{+}) / \tau)}
       {\exp(s(q, \hrepo{+}) / \tau) + \sum_{\hrepo{-}} \exp(s(q, \hrepo{-}) / \tau)},
\end{equation}
where $s(\cdot,\cdot)$ is cosine similarity and $\tau{=}0.07$.

\paragraph{Separate regime.} After contrastive pretraining, we freeze $F_\theta$ and train only $\pi_\psi$ and $G_\phi$ with the causal language modelling loss:
\begin{equation}
  \mathcal{L}_{\mathrm{CLM}} = -\sum_{t=1}^{|y^*|} \log p_\phi(y^*_t \mid y^*_{<t}, x, \mathcal{P}).
\end{equation}

\paragraph{End-to-end regime.} We initialise from the contrastive checkpoint and train $F_\theta$, $\pi_\psi$, and $G_\phi$ jointly with $\mathcal{L}_{\mathrm{CLM}}$. Gradients propagate through the entire pipeline except the frozen embedder $E$. This yields the best accuracy.

\subsection{Utility-Weighted Likelihood (UWL) for Data Filtering}
\label{sec:uwl}

We use \textbf{Utility-Weighted Likelihood}~\cite{ghossein2024iclerb} to filter training data, keeping only contexts that improve completion quality.

\paragraph{Definition.} For prefix $x$, candidate context chunk $c$ (raw text), and gold completion $y^*$, we compute the log-likelihood gain under a frozen reference model $M$:
\begin{equation}
  \Delta(x, c) = \log p_M(y^* \mid x, c) - \log p_M(y^* \mid x).
  \label{eq:uwl}
\end{equation}
Positive $\Delta$ means context $c$ raises the likelihood of the correct completion.

\paragraph{Filtering.} We apply sigmoid and threshold: $\text{UWL}(x,c) = \sigma(\Delta(x,c))$. Training pairs are kept only if $\text{UWL}(x,c) \ge 0.55$. UWL is used \emph{only} for offline data filtering, not at inference.

\subsection{Extensions}
\label{sec:extensions}

We describe two optional enhancements.

\paragraph{Entity pre-parsing.} Error analysis revealed that many failures stem from hallucinated identifiers: the decoder cannot reconstruct rare symbol names from fused vectors alone. We run an AST walk to extract fully qualified names of user-defined entities and append them as up to 64 extra tokens. Section~\ref{sec:ablation} quantifies the improvement.

\paragraph{Per-chunk p-tuning.} The chief bottleneck of the HEF pipeline is static chunk embeddings: they capture coarse semantic similarity but discard fine-grained lexical detail needed for precise identifier reconstruction. We borrow per-sample prompt-tuning from~\citet{kuratov2025cramming1568tokenssingle}, who showed that a single trainable vector can losslessly store 1568 tokens with compression ratio up to $\times$1500. Concretely, for every code chunk we attach one learnable \texttt{[mem]} token and optimize it with causal LM loss (next-token prediction on the chunk itself) for 250--500 AdamW steps against a frozen \texttt{Qwen-2.5-Coder-0.5B} backbone. Despite relying on the small 0.5B model, the p-tuned vectors lift exact-match accuracy. The downside is steep indexing cost: because each chunk is optimized independently, preprocessing time rises from minutes to hours per repository on a single A100.

\begin{table*}[t]
  \caption{Repository-level code-completion results. Exact-match percentages ($\uparrow$). Latency measured on single A100.}
  \label{tab:hef_main_results}
  \centering
  \begin{tabular}{l c c c c c}
    \toprule
    \textbf{Approach} & \textbf{Params} & \textbf{Latency} & \textbf{Offline} & \textbf{RepoEval} & \textbf{RepoBench} \\
    \midrule
    \multicolumn{6}{c}{\textit{High-latency models}}\\
    \midrule
    DRACO \cite{cheng2024draco}         & 7.1B & 11.0s & 14.3s  & 46.4 & 60.3 \\
    GraphCoder \cite{liu2024graphcoder} & 16B  & 17.5s & 10.2s  & 48.7 & 64.1 \\
    RLCoder \cite{wang2024rlcoder}      & 1.1B &  8.8s & 50.4s  & 49.9 & 65.9 \\
    \midrule
    \multicolumn{6}{c}{\textit{Low-latency models}}\\
    \midrule
    Qwen-2.5-Coder-1.3B           & 1.5B & 0.50s &    {-} & 28.4 & 49.1 \\
    RepoFusion \cite{shrivastava2023repofusion} & 0.22B & 0.9s &  24.6s & 33.2 & 39.8 \\
    \textbf{HEF (separate)}                  & 1.8B & 0.94s &  35.1s & 34.1 & 56.9 \\
    \textbf{HEF (end-to-end)}                & 1.8B & \textbf{0.68s} & 35.1s & 42.7 & \textbf{61.3} \\
    \textbf{HEF (e2e + p-tuning)}      & 1.8B & 0.59s & 1806s & \textbf{43.9} & N/A\footnotemark \\
    \bottomrule
  \end{tabular}
\end{table*}
\footnotetext{Per-chunk p-tuning is computationally intensive; RepoBench was not evaluated in this setup.}

\section{Experimental Setup}
\label{sec:setup}

\subsection{Baselines}
We compare our model to three current so-called \textit{high-latency baselines}, i.e. models with latency significantly higher than 1 second, namely: DRACO, GraphCoder, and RLCoder.

\textbf{DRACO}~\cite{cheng2024draco} represents each project as an explicit data- and control-flow graph, then walks that graph at inference time to surface semantically related code snippets.

\textbf{GraphCoder}~\cite{liu2024graphcoder} builds a coarser context graph and applies a coarse-to-fine reranking strategy to pick the most relevant nodes. Both methods therefore depend on reparsing the repository and performing graph traversals whenever a prediction is requested, which creates noticeable latency and memory overhead on large or frequently changing codebases.

\textbf{RLCoder}~\cite{wang2024rlcoder} formulates retrieval as a reinforcement-learning problem: a dual-encoder iteratively adds chunks until the estimated utility stalls.

We also compare to a representative \textit{low-latency baseline}:
\textbf{RepoFusion}~\cite{shrivastava2023repofusion}, which concatenates the full repository into the decoder's context during pre-training, eliminating explicit retrieval at inference time.

\subsection{Training Data}
We continuously mirror high-quality public repositories ($\mathrm{stars} > 500$) from GitHub. Repositories are deduplicated at the file level and filtered by permissive licenses (MIT, Apache-2.0, BSD). After deduplication and filtering, our training corpus contains approximately 120K Python and 45K Java repositories.

\subsection{Implementation Details}
\label{sec:impl}

Table~\ref{tab:impl_details} summarises the key hyperparameters and model configurations.

\begin{table}[t]
\caption{Implementation details.}
\label{tab:impl_details}
\small
\begin{tabular}{l l}
\toprule
\textbf{Component} & \textbf{Configuration} \\
\midrule
\multicolumn{2}{l}{\textit{Model architectures}} \\
Embedder $E$ & Qwen3-Embedding-8B, $d{=}4096$ \\
Fuser $F_\theta$ & Qwen-2.5-Coder-0.5B, $d_f{=}896$ \\
Generator $G_\phi$ & Qwen-2.5-Coder-1.5B, $d_g{=}1536$ \\
Projector $\pi_\psi$ & 2-layer MLP, hidden 2048 \\
\midrule
\multicolumn{2}{l}{\textit{Training hyperparameters}} \\
Optimizer & AdamW ($\beta_1{=}0.9$, $\beta_2{=}0.999$) \\
Learning rate & $2 \times 10^{-5}$ (cosine decay) \\
Warmup steps & 500 \\
Batch size & 32 (effective, with grad.\ accum.) \\
Contrastive steps & 10K \\
End-to-end steps & 50K \\
\midrule
\multicolumn{2}{l}{\textit{Pipeline settings}} \\
Chunk size & $\le$512 tokens \\
Branching factor $b$ & 8 \\
Retrieved nodes $K$ & 32 \\
UWL threshold $\tau$ & 0.55 \\
\midrule
\multicolumn{2}{l}{\textit{Hardware}} \\
Training & 4$\times$A100 40GB \\
Inference & 1$\times$A100 40GB \\
\bottomrule
\end{tabular}
\end{table}

\paragraph{Reproducibility.} We will publicly release the training scripts, data preparation pipeline, and model checkpoints upon publication to facilitate reproducibility and future research.

\subsection{Datasets}
We benchmark HEF on two public repository-level suites, RepoBench~\cite{liu-etal-2024-repobench} and RepoEval~\cite{zhang-etal-2023-repoeval}, which together provide more than 62k completion targets.

\textbf{RepoBench v1.0}~\cite{liu-etal-2024-repobench} consists of 1,075 Python and 594 Java test repositories collected after February 2023 to avoid overlap with common training corpora. We evaluate all approaches on the RepoBench-C sub-task (code completion), which contains 61k next-line examples per language under 2k- and 8k-token context budgets. The evaluation metric is Exact-Match (EM) accuracy.

\textbf{RepoEval}~\cite{zhang-etal-2023-repoeval} comprises 14 high-quality GitHub projects created in 2022--2023. We evaluate on the Line Completion task, which contains 1,600 masked lines (200 per repository). While the benchmark also defines Edit Similarity (ES) metrics for line and API tasks, we report Exact-Match (EM) accuracy for consistency with RepoBench.

During training, we rigorously verified that none of the repositories from these benchmarks, nor their forks, were included in our training data.

\section{Results}
\label{sec:results}

Table~\ref{tab:hef_main_results} summarises the numbers for HEF and a
set of high- and low-latency baselines that have
comparable public checkpoints.

\subsection{Overall Accuracy}

End-to-end HEF reaches \textbf{61.3\%} exact match on
RepoBench, 12.2 points above the plain
Qwen-2.5-Coder-1.3B baseline and 22.5 points above the low-latency baseline RepoFusion.
It is also 4.4 points above the
contrastively trained (``separate'') HEF variant.

On RepoEval the margin is wider for
the baseline, namely 42.7\% versus 28.4\%. For RepoFusion it is more than 10 points.

These gains close much of the gap to far larger systems---GraphCoder runs a 16B decoder yet scores only 64.1\% on
RepoBench, a lead of 2.8 points while using over eight times
as many parameters in the language model alone and being 20 times slower.

\subsection{Efficiency}

Despite the accuracy boost, HEF keeps inference time low. The median
latency of 0.68s makes it roughly 26$\times$ faster than GraphCoder
(17.5s) and 13$\times$ faster than DRACO (11.0s). Classical retrieval
pipelines have to inject thousands of raw tokens; HEF transmits only a
few dozen pseudo-tokens plus a single forward pass through the
0.5B fuser and 1.3B-parameter decoder. Offline preprocessing---the one-time cost of
building repository embeddings---takes 35s per project on an A100, which
is well below the 50s required by RLCoder's reinforcement-learning
stage.

\subsection{Separate versus End-to-End Training}

Training the fuser with a contrastive objective already lifts the
baseline by 7.8 points on RepoBench. Allowing gradients to
flow through the fuser and decoder jointly brings another 4--9 point
gain across the two benchmarks, suggesting that tight alignment
between the hierarchical vectors and token probabilities is worth the
extra compute during training.

\subsection{Comparison to Prior Work}

Graph- and reinforcement-learning-based retrievers such as DRACO,
GraphCoder and RLCoder improve relevance at the cost of multiple model
calls or heavy graph traversals. HEF attains similar or better
accuracy with just one encoder pass per repository and one decoder
pass per query, avoiding per-query structural analysis altogether.
RepoFusion trades retrieval for brute-force context expansion, pushing
the entire repository into the decoder every time; it achieves only
39.8\% on RepoBench, far below either HEF variant, while
still paying a higher latency than the end-to-end model.

Hierarchical fusion lets a modest 1.8B-parameter system compete with
or surpass methods that use larger decoders, complex graph
inference or reinforcement learning, all while answering in well under
a second. The results demonstrate that most repository-level
information can be distilled into a compact vector hierarchy without
sacrificing quality.

\begin{table}[t]
\caption{Overall ablation on the validation set. All numbers are average \textsc{CodeBLEU}.}
\label{tab:ablation_overall}
\begin{tabular}{l r}
\toprule
\textbf{Variant} & \textbf{CodeBLEU} \\
\midrule
Initial model & 23.32 \\
HEF (separate) & 32.31 \\
HEF (end-to-end, random ctx) & 24.53 \\
HEF (end-to-end, UWL ctx) & 36.20 \\
HEF (end-to-end, UWL ctx, 52 tokens) & 38.35 \\
HEF (e2e, UWL ctx, 52 tok, entity parsing) & \textbf{42.80}\\
\bottomrule
\end{tabular}
\end{table}

\begin{figure}[t]
\centering
\includegraphics[width=0.95\columnwidth]{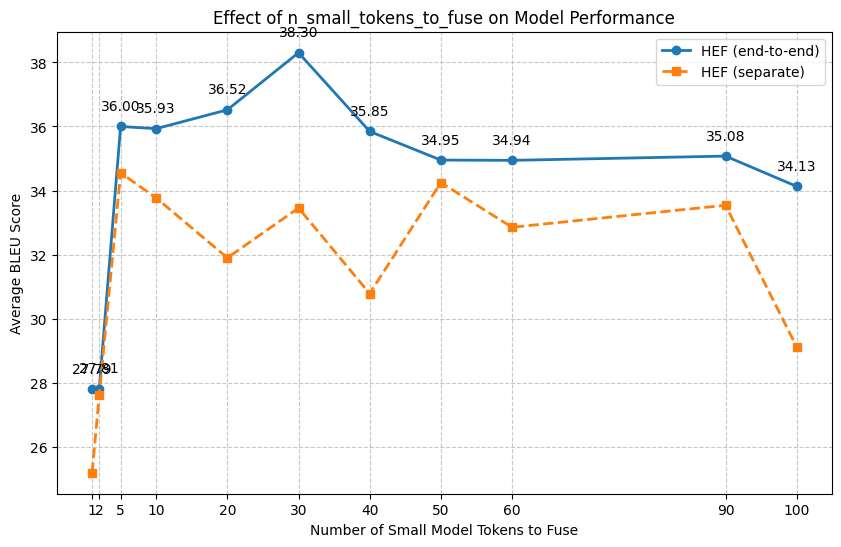}
\caption{Effect of the number of fused embedding tokens on \textsc{CodeBLEU}. Labels show exact scores at each point.}
\Description{A line graph showing CodeBLEU scores increasing from 1 token to around 30-40 tokens, then slightly declining after 60 tokens.}
\label{fig:fuse_tokens_bleu}
\end{figure}

\section{Ablation Study}
\label{sec:ablation}

We conduct a series of controlled experiments to disentangle the impact of each design choice in Hierarchical Embedding Fusion (HEF). Specifically, we vary
(1) the training regime of the Fuser (random vs. UWL-filtered context),
(2) whether the Fuser is optimised separately or end-to-end with the generator, and
(3) the number of small-model embedding tokens that are injected into the prompt at inference time.

\subsection{How Many Fused Tokens Do We Need?}
\label{subsec:tokens_to_fuse}

Figure~\ref{fig:fuse_tokens_bleu} plots the average CodeBLEU on the held-out set as we progressively increase the number of fused tokens. For HEF (end-to-end), the score surges from 27.8 to 36.0 when moving from 1 to 5 tokens, peaks at 38.3 with $n=30$, and slowly declines afterwards. HEF (separate) follows a similar trend but tops out at 35.0. These results suggest that roughly 30--40 fused tokens already capture most repository-level information; pushing beyond 60 brings diminishing---or even negative---returns, presumably because the decoder's fixed capacity is wasted on redundant vectors.

\subsection{Training Regime and Data Quality}

Table~\ref{tab:ablation_overall} contrasts five variants. Training the Fuser separately already lifts the baseline by +9.0 points. End-to-end optimisation with randomly sampled context barely helps (+1.2), indicating that gradient signal is weak when most retrieved chunks are irrelevant. Switching to UWL-filtered context, however, yields a sizable jump to 36.2 (+12.9). Finally, applying the optimal setting found in Section~\ref{subsec:tokens_to_fuse} (52 fused tokens) pushes the score to 38.35, our best result.

\subsection{Robustness to Irrelevant or Harmful Context}
Retrieval systems occasionally surface misleading snippets. To stress-test robustness, we isolate examples whose UWL score is negative (i.e., the retrieved chunk hurts the likelihood of the reference answer). The results are shown in Table~\ref{tab:dpo_negative}. The vanilla \textsc{Qwen-2.5-Coder-1.5B-Instruct} model reaches 18.10 \textsc{CodeBLEU}. Naively appending the top 4k raw tokens degrades performance to 14.40 (-3.7). In contrast, HEF (which distils the same context into a compact embedding tree) retains a higher 16.70. Although absolute scores remain low, HEF cuts the relative error increase by more than one third, confirming that the fusion mechanism is \emph{less sensitive} to noise than plain RAG.

\begin{table}[t]
\caption{Evaluation on the subset where UWL scorer predicts negative utility for all retrieved chunks.}
\label{tab:dpo_negative}
\begin{tabular}{l r}
\toprule
\textbf{Model / Context Strategy} & \textbf{CodeBLEU} \\
\midrule
Qwen-2.5-Coder-1.5B-Instruct & 18.10 \\
+ 4k raw tokens & 14.40 \\
HEF & 16.70 \\
\bottomrule
\end{tabular}
\end{table}

\subsection{Effect of the Embedding Backbone}
\label{sec:embedding_ablation}

Table~\ref{tab:embedding_ablation} extends our earlier analysis with the new
p-tuned variant that optimises a single [mem] token for each chunk on
top of the small Qwen-2.5-coder-0.5B.
Although the underlying backbone is the lightest in the pool, the extra
fine-tuning lifts RepoEval accuracy to 43.9\%,
overtaking even the 8-billion-parameter model.

\begin{table}[t]
\caption{Embedding-model ablation ($\uparrow$ higher is better).}
\label{tab:embedding_ablation}
\begin{tabular}{lccc}
\toprule
\textbf{Model} & \textbf{RepoEval} & \textbf{MTEB} & \textbf{Dim.}\\
\midrule
GTE-7B                              & 39.5 & 70.7 & 3584 \\
Qwen3-Embedding-0.6B                & 36.7 & 70.7 & 1024 \\
Qwen3-Embedding-4B                  & 38.2 & 74.6 & 2560 \\
Qwen3-Embedding-8B                  & 42.7 & 75.2 & 4096 \\
Qwen-2.5-coder-0.5B + p-tuning      & \textbf{43.9} & n/a & 896 \\
\bottomrule
\end{tabular}
\end{table}

\subsection{Effect of Fuser Model Size}
\label{sec:fuser_size_ablation}

Table~\ref{tab:fuser_size} examines the impact of fuser capacity on downstream accuracy and cache construction cost. Scaling the fuser from 0.5B to 1.5B parameters yields only a modest +0.8 CodeBLEU improvement, while tripling offline build time. Larger fusers (3B) provide no additional gain and further increase indexing cost. These results confirm that the fuser's role is compression rather than reasoning: a compact model suffices to aggregate child vectors, and the quality bottleneck lies in the downstream generator.

\begin{table}[t]
\caption{Fuser size ablation on the validation set.}
\label{tab:fuser_size}
\begin{tabular}{lccc}
\toprule
\textbf{Fuser} & \textbf{CodeBLEU} & \textbf{Offline (s)} & \textbf{Params}\\
\midrule
Qwen-2.5-Coder-0.5B  & \textbf{38.35} & \textbf{35.1} & 0.5B \\
Qwen-2.5-Coder-1.5B  & 39.12 & 98.4 & 1.5B \\
Qwen-2.5-Coder-3B    & 39.08 & 187.2 & 3B \\
\bottomrule
\end{tabular}
\end{table}

\section{Conclusion}
\label{sec:conclusion}

We studied repository-level code completion under interactive latency constraints and introduced Hierarchical Embedding Fusion (HEF), a pipeline that amortizes repository processing by caching a project as a hierarchical dense representation and conditioning generation through a fixed pseudo-token interface. By decoupling online prompt length from repository size, HEF enables sub-second end-to-end completion while preserving most of the accuracy of snippet-based retrieval methods.

HEF is not designed to replace high-latency graph or iterative-retrieval systems in accuracy-critical settings. Rather, it occupies a complementary region of the accuracy-latency tradeoff, offering a practical alternative when responsiveness is a primary concern. Our results on RepoBench and RepoEval show that compact fused representations can retain repository-level cues relevant to completion while dramatically reducing online computation.

Beyond individual components, our contribution is an end-to-end recipe for repository-aware completion: a hierarchical dense cache built from a strong embedder, a lightweight fusion model, a pseudo-token conditioning interface, and an unsupervised data construction procedure. Together, these elements demonstrate that repository context can be integrated without streaming large volumes of raw code into the generator.

We see several promising directions for future work, including adaptive hierarchy construction, level-aware retrieval policies, and hybrid methods that combine symbolic program structure with compact continuous caches. We hope this work encourages further exploration of low-latency, repository-aware generation beyond raw context injection.

\begin{acks}
To be added.
\end{acks}

\bibliographystyle{ACM-Reference-Format}
\bibliography{lit}

\end{document}